\documentclass[conference]{IEEEtran}
\IEEEoverridecommandlockouts

\usepackage{cite}
\usepackage{amsmath,amssymb,amsfonts}
\usepackage{algorithmic}
\usepackage{graphicx}
\usepackage{textcomp}
\def\BibTeX{{\rm B\kern-.05em{\sc i\kern-.025em b}\kern-.08em
    T\kern-.1667em\lower.7ex\hbox{E}\kern-.125emX}}

\usepackage{amsmath,amssymb,amsfonts}
\usepackage{algorithmic}
\usepackage{graphicx}
\usepackage{textcomp}
\usepackage{multirow}
\usepackage[table,xcdraw]{xcolor}
\usepackage[normalem]{ulem}
\useunder{\uline}{\ul}{}
\usepackage{url}
\usepackage{subcaption}

\begin{document}
\title{
Efficient Training for Compact Compression Models via Sequential Distillation
\\
\thanks{The authors acknowledge fundings of France 2030, PEPR IA, ANR-23-PEIA-0008 and European Union ERC-2024-STG-101163069 MALAGA.}
}

\author{
\IEEEauthorblockN{Caroline Mazini Rodrigues,Nicolas Keriven,Thomas Maugey}
\IEEEauthorblockA{Univ. Rennes, Inria, CNRS, IRISA, Rennes, France}
}

\maketitle

\begin{abstract}
Deep learning models for image compression often face practical limitations in hardware-constrained applications. Although these models achieve high-quality reconstructions, they are typically complex, heavyweight, and require substantial training data and computational resources. We propose a methodology to significantly reduce autoencoder-based compression networks in a more stable Knowledge Distillation process. The intuition is that highly reduced architectures benefit from simplified optimization objectives in early training, with complexity gradually introduced later. Therefore, our approach begins with a sequential encoder–decoder distillation stage that provides a robust initialization for the lightweight model. This is followed by standard training that can be regularized with latent distillation. We evaluate the resulting lightweight autoencoders across two different architectures on the image compression task. Experiments show that our method preserves reconstruction quality and statistical fidelity in early epochs better than training lightweight autoencoders with the original loss, making it practical for resource-limited environments.
\end{abstract}


\section{Introduction}


Deep learning models have increasingly been adopted in image compression applications where handcrafted codecs~\cite{wallace:tce:1992:jpeg, lainema:tcsvt:2012:HEVC/VVC,pfaff:tcsvt:2021:vvc} were once more common~\cite{yang:2023:trends:intro1}. This shift is due not only to their ability to approximate complex functions, but also to their capacity to learn from and adapt to large-scale data. Compared to traditional codecs, end-to-end Learned Image Coders (LICs) such as the ones using encoder-decoder architectures~\cite{balle:pcs:2016:compr:structure,balle:iclr:2017:factorized,balle:iclr:2018:hyper}, are larger models that require more computational resources, as well as more time and data for training, which can limit their practical use. In streaming applications, for instance, it is preferable to use lightweight decoders that can run efficiently on general-purpose hardware, such as mobile devices~\cite{yang:iccv:2023:shallow:decoder}. In contrast, within the Internet-of-Things (IoT) context --- where the use of connected devices continues to grow --- it is even more important to deploy compact encoders~\cite{hojjat:arxiv:2024, Gilbert:2024:ahn, zheng:arxiv:2024}, allowing encoding on the hardware-limited devices and simplifying data transmission~\cite{liang:2020:itj}.

Although there is no clear consensus on the necessary size of a network architecture to approximate a given function, some studies suggest that smaller subparts of the models can achieve comparable performance~\cite{frackle:iclr:2019:lotery:ticket}. However, simply reducing the size of deep learning models does not necessarily preserve their performance. Studies such as that by Duan et al.~\cite{Duan:2023:icip} show that deeper and larger networks can help narrow the gap to the optimal solution in terms of rate-distortion trade-offs. 

Given large, well-trained models that have learned robust feature representations, we aim to use their knowledge to guide smaller models in obtaining comparable performance. One established approach for transferring knowledge from large deep learning models,  trained on massive datasets, to smaller models is \textit{Knowledge Distillation} (KD)~\cite{Hinton:arxiv:2015:DistillingTK, liu:grsl:2024:SAR}, which adds an extra optimization objective: encouraging the small model (student) to produce features or outputs similar to those of the larger model (teacher). However, general-purpose KD methods do not always account for time and hardware constraints, as they can increase optimization complexity. 

In this paper, we address the problem of reducing the model size of learned image compression (LIC) models under limited computational resources, reflected as limited training epochs. \textit{We hypothesize that, to avoid instabilities during limited training of significantly reduced architectures, it is necessary to simplify the objective, at least in the early epochs.} \textit{By relying exclusively on KD guidance from larger models, we can induce this simplification during early training.} To test this hypothesis, we introduce a preliminary sequential encoder–decoder distillation strategy adapted to the compression pipeline. At first, instead of transferring all knowledge at once, we split the distillation process into two sequential phases to gradually reduce model complexity. A compact student encoder is first trained to mimic the latent representations of a larger teacher encoder over a few epochs. Next, a compact decoder is trained to reconstruct the input from the student's latent space, completing the reduction of model size. Experiments showed that this preliminary strategy improves performance compared to other reduction strategies in early training epochs.

\section{Autoencoder reduction}

\textbf{Original Architecture:} We employ autoencoder architectures, as they are among the most used models for lossy image compression tasks~\cite{Jamil:2023:EAAI}. Within this category of architectures, we analyze two types of models. The first is a more general approach that uses a single autoencoder for compression, known as the \textit{Factorized Prior} ($FP$) model~\cite{balle:iclr:2017:factorized}. The second employs two autoencoders --- one for compression and another to learn a prior --- named \textit{Hyperprior} ($HP$) model~\cite{balle:iclr:2018:hyper}. 

Both models are trained with a Rate-Distortion (RD) loss, minimizing the bit-rate of the \textbf{latent representation} $y = g_a(x)$ while preserving similarity between the original image $x$ and its reconstruction $\hat{x} = g_s(g_a(x))$, thereby minimizing distortion. In this setting, the encoder is denoted by $g_a(\cdot)$, and the decoder by $g_s(\cdot)$. The RD loss is defined in Equation~\eqref{eq:rd_loss}:
\begin{equation}
    \mathcal{L}_{RD}(x,\hat{x},\hat{y}) =  -\log_2 p_{\hat{y}}(\hat{y}) +  \lambda \times ||x- \hat{x}||_2^2,
    \label{eq:rd_loss}
\end{equation}
where the first term penalizes bit-rate, the second term reconstruction error, $\hat{y} = q(y)$ is the quantized latent representation, $p_{\hat{y}}(\hat{y})$ is learnt by Entropy Bottleneck $EB(\hat{y})$ (a probabilistic model), and $\lambda$ controls the trade-off rate-distortion. 

While the $FP$ model assumes that the latents $y$ are independent, $HP$ models capture dependencies among the latents through a conditional distribution. In $HP$, a second autoencoder processes the latents: the hyper-encoder produces hyper-latents that are also quantized $\hat{z} = q(h_a(\hat{y}))$, and the hyper-decoder outputs side information $\tilde{y} = h_s(\hat{z})$ that conditions the entropy model $EB(\hat{y})$. This results in a conditional prior $p_{\hat{y}\mid \hat{z}}(\hat{y} \mid \hat{z})$, and the rate-distortion loss is:
\begin{equation}
\mathcal{L}_{RD}(x, \hat{x}, \hat{y}, \hat{z}) = -\log_2 p_{\hat{y}\mid \hat{z}}(\hat{y} \mid \hat{z}) - \log_2 p_{\hat{z}}(\hat{z}) + \lambda \, \|x - \hat{x}\|_2^2.
\label{eq:rd_loss_hyper}
\end{equation}

These are large architectures trained on big datasets, such as Vimeo-90k (89,800 videos) and OpenImagesV6 (9M images). 

\textbf{Reduction objectives:} We want to exploit the knowledge from complex pretrained models $\mathcal{T}$ to help to obtain lightweight autoencoder versions of the same architecture, $\mathcal{S}$, that achieve comparable rate-distortion performance.
We target the reduction of two main aspects in autoencoder-based compression networks: the \textit{size of models} and the \textit{complexity of training} (here, number of epochs).

\textit{1. Model reduction:} Our goal is to reduce the models' size while maintaining similar rate-distortion performance. The reduction is performed by scaling down the widths of the convolutional layers in encoder/decoder by a factor of $r$. 

\textit{2. Training complexity reduction:} Modern image datasets used to train compression models are large, making training computationally expensive and sensitive to instability. Models such as $FP$ are typically trained for 4-5M steps~\cite{begaint:2020:compressai}, corresponding to hundreds of epochs, depending on the dataset size. We aim at proposing more stable training strategies to \emph{significantly reduce the number of training epochs} while maintaining competitive performance, thereby lowering training costs. 
\section{Distillation strategy}

In this paper, we explore the use of Knowledge Distillation (KD) to efficiently reduce image compression autoencoders. Training much smaller models using only the standard loss fails to achieve the intended results. KD~\cite{romero:arxiv:2015:fitnetshintsdeepnets} is a method that transfers knowledge from a model $\mathcal{T}$, called the \textit{teacher}, to another model $\mathcal{S}$, called the \textit{student}. It is commonly used to compress large models into smaller ones. In KD, the student is trained using both the original task loss (Equations~\eqref{eq:rd_loss} and~\eqref{eq:rd_loss_hyper}) and an additional distillation loss. This additional loss encourages the student to behave like the teacher, either by matching the teacher's intermediate features or their final outputs. However, combining multiple loss terms increases training complexity. This can make optimization more expensive and less stable, especially when the student model is much smaller than the teacher.

Prior work~\cite{Chen:2025:iccv} primarily focuses on distilling knowledge to replicate the rate–distortion behavior of a teacher model, without considering the cost of the distillation process or different model reduction rates. In contrast, we aim to avoid excessive training by adopting a more stable distillation strategy and by explicitly evaluating different architecture reduction rates.

\textit{Our intuition is that starting with a simpler objective allows us to smoothly introduce complexity, maintaining training stability and achieving better performance within few epochs.}  

\textbf{Preliminary sequential encoder-decoder distillation:} 
We propose to initially train the two modules, encoder and decoder, separately and sequentially, gradually reducing the autoencoder's size (to prevent aggressive reductions), as shown in Fig.~\ref{fig:enc_reduction}, with each module guided by specific parts of the teacher’s knowledge. This allows us to simplify the training loss for each module. The process is described as follows:

    \begin{figure}[!ht]
    \centerline{\includegraphics[width=\linewidth]{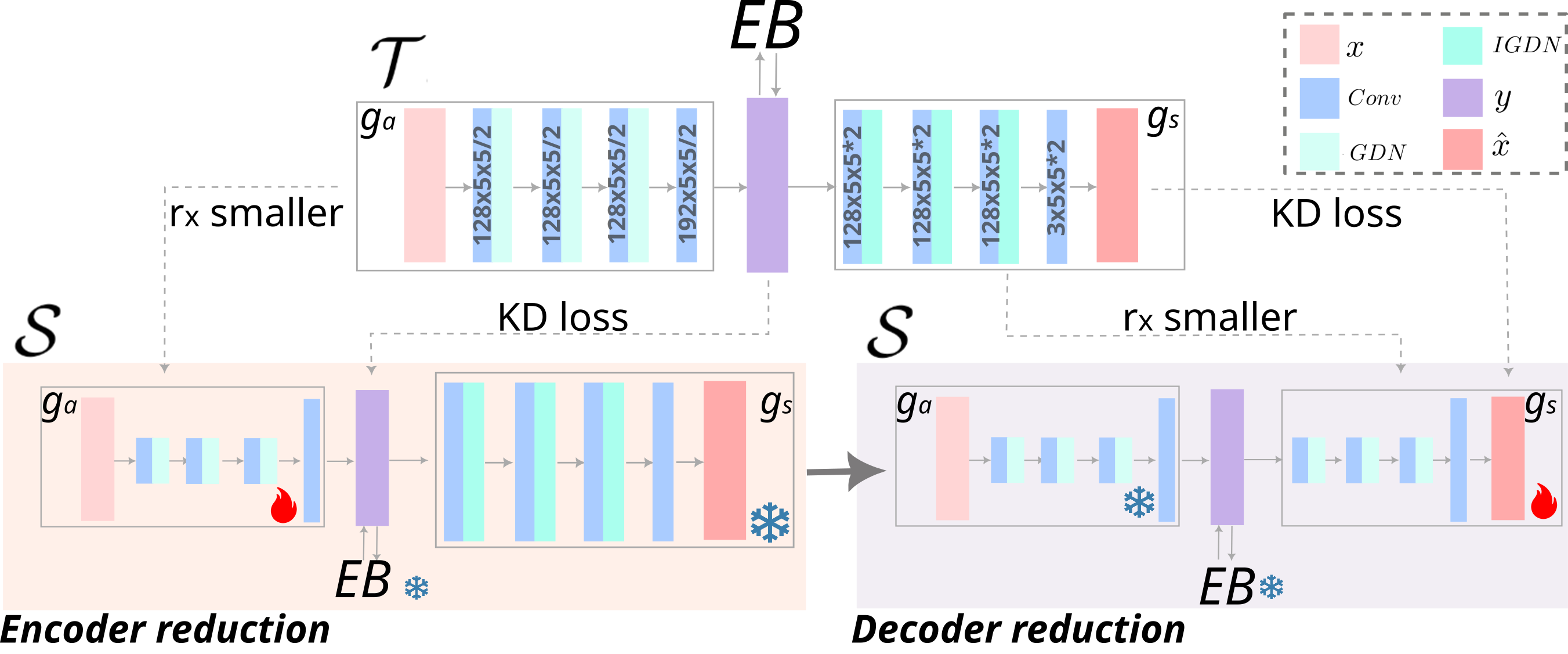}}
    \caption{\small{\textit{Original ($\mathcal{T}$) and reduced ($\mathcal{S}$) Factorized prior architectures.} The fire icon denotes the trainable part, while the snowflake indicates the frozen part. The Orange block is $i$) Encoder reduction, and the purple block is $ii$) Decoder reduction, during preliminary distillation.}}
    \label{fig:enc_reduction}
    \end{figure}
    
\textit{i) Encoder reduction.}  
    We train a reduced encoder by a factor $r$. The decoder and entropy bottleneck are \emph{fixed}, shared with the teacher model, \textit{i.e.},
    $g_s^{\mathcal{S}}(\cdot) = g_s^{\mathcal{T}}(\cdot)$ and
    $EB^{\mathcal{S}}(\cdot) = EB^{\mathcal{T}}(\cdot)$.
    In the  $HP$ case, we additionally set
    $h_s^{\mathcal{S}}(\cdot) = h_s^{\mathcal{T}}(\cdot)$.
    These modules are frozen, while only the reduced encoder components,
    $g_a^{\mathcal{S}}(\cdot)$ and (when applicable) $h_a^{\mathcal{S}}(\cdot)$, are trained. In this phase, we modify the loss and \emph{only train the student encoder to match the teacher's latent representations $y_\mathcal{T}$}.
This helps the encoder student focus on a single task --- matching the teacher's latent representations --- while also keeping the student’s compression rate under control. 
For the $FP$ architecture, we minimize only a KD loss from Equation~\eqref{eq:loss_KD_alone}: 
\begin{equation}
\mathcal{L}_{KD}^{enc}(y_\mathcal{T}, y_\mathcal{S}) = \|y_\mathcal{T} - y_\mathcal{S}\|_2^2.
 \label{eq:loss_KD_alone}
\end{equation}

For 
the $HP$ model, the latents from both reduced modules, encoder and hyper-encoder, $y_\mathcal{T}$ and $z_\mathcal{T}$ are approximated, resulting in the loss function defined in Equation~\eqref{eq:loss_hyper_KD_alone}: 
\begin{equation}
\mathcal{L}_{KD}^{enc}(y_\mathcal{T},z_\mathcal{T}, y_\mathcal{S}, z_\mathcal{S}) = \|y_\mathcal{T} - y_\mathcal{S}\|_2^2 + \|z_\mathcal{T}- z_\mathcal{S}\|_2^2.
 \label{eq:loss_hyper_KD_alone}
\end{equation}

\textit{ii) Decoder reduction.}  
    In the second stage, with the smaller version of the encoder optimized to replicate the teacher's latents, we train the decoder (reduced by the same factor $r$). The previously reduced encoder,
    $g_a^{\mathcal{S}}(\cdot)$ (and $h_a^{\mathcal{S}}(\cdot)$ in the Hyperprior case), is frozen, and only the reduced decoder components,
    $g_s^{\mathcal{S}}(\cdot)$ (and $h_s^{\mathcal{S}}(\cdot)$), are optimized. In this phase, \emph{we train the student decoder to match the teacher's reconstruction $\hat{x}_\mathcal{T}$}, while keeping the (already reduced) student encoder frozen. For $FP$, since the compression rate is already determined by the approximated latents, the decoder is trained only using the reconstruction-based distillation loss in Equation~\eqref{eq:loss_KD_alone_dec}: 
\begin{equation}
\mathcal{L}_{KD}^{dec}(\hat{x}_\mathcal{T}, \hat{x}_\mathcal{S}) = \|\hat{x}_\mathcal{T} - \hat{x}_\mathcal{S}\|_2^2.
 \label{eq:loss_KD_alone_dec}
\end{equation}

For the $HP$ architecture, with the extra hyper-decoder module $h_s(\cdot)$, the training requires regularization to ensure valid entropy modeling. We therefore include the original rate terms in the decoder loss, resulting in Equation~\eqref{eq:loss_hyper_KD_alone_dec}: 
\begin{equation}
\begin{aligned}
\mathcal{L}_{KD}^{dec}(\hat{x}_\mathcal{T}, \hat{x}_\mathcal{S}, \hat{y}_\mathcal{S}, \hat{z}_\mathcal{S})
&= -\log_2 p_{\hat{y}_\mathcal{S}\mid \hat{z}_\mathcal{S}}(\hat{y}_\mathcal{S} \mid \hat{z}_\mathcal{S})
   - \log_2 p_{\hat{z}_\mathcal{S}}(\hat{z}_\mathcal{S}) \\
&\quad + \lambda \, \|\hat{x}_\mathcal{T} - \hat{x}_\mathcal{S}\|_2^2 .
\end{aligned}
\label{eq:loss_hyper_KD_alone_dec}
\end{equation}

\textbf{Latent distillation as regularization:} After this initial phase of sequential encoder–decoder distillation, the smaller model is warmed up for the regular training (original loss). To prevent forgetting the knowledge learned from the teacher, we keep using KD as a regularization term to encourage the student's latent representations to remain similar to those of the teacher. The final loss function is given in Equation~\eqref{eq:rd_loss_regularized}:
\begin{equation}
    \mathcal{L}_{final}(\cdot) =  \alpha \times \mathcal{L}_{RD}(\cdot) + \beta \times \mathcal{L}_{KD}^{enc}(\cdot)
    \label{eq:rd_loss_regularized}
\end{equation}
with $0<\alpha< 1$, $0<\beta < 1$ and loss parameters depending on the architecture: for $FP$, as defined in Equations~\ref{eq:rd_loss} and \ref{eq:loss_KD_alone}, and for $HP$, as defined in Equations~\ref{eq:rd_loss_hyper} and \ref{eq:loss_hyper_KD_alone}. While $\alpha$ and $\beta$ can be adaptive during training, we used a fixed weighting (for $FP$ $\beta = 10 \times \alpha$, for MS-ILLM $\beta = \alpha$).

\begin{figure*}[!ht]
\centering
 \begin{subfigure}[t]{0.3\textwidth}
\centering
\includegraphics[width=\textwidth]{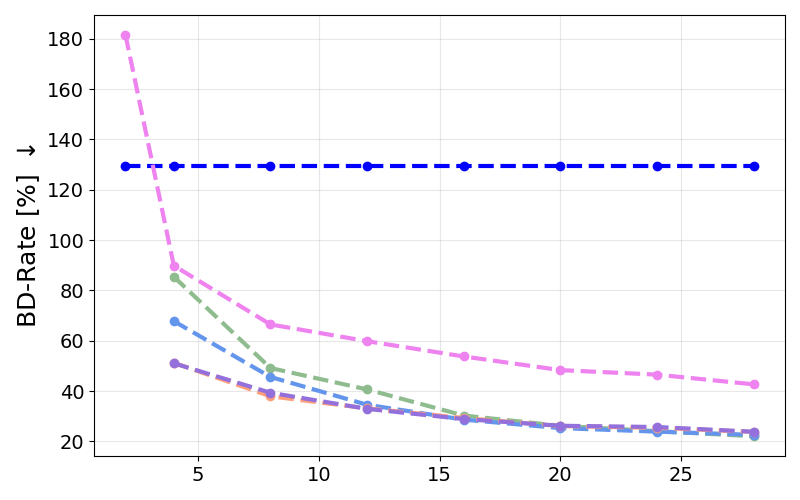}
    \end{subfigure}
~
    \begin{subfigure}[t]{0.3\textwidth}
        \centering
\includegraphics[width=\textwidth]{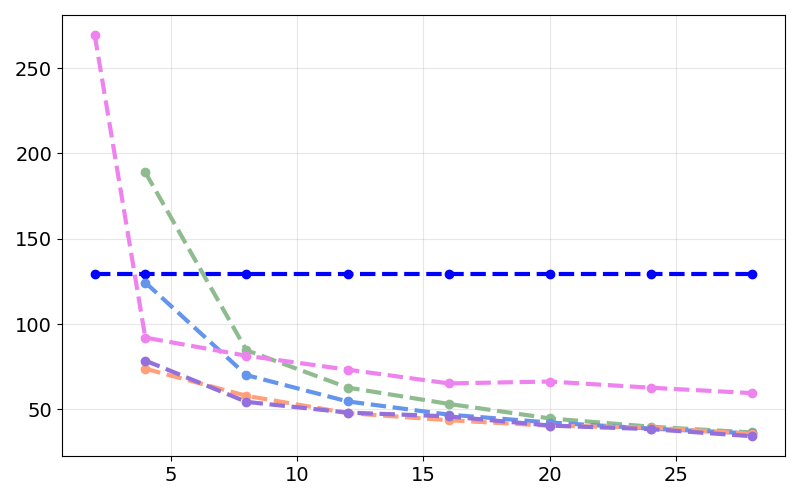}
    \end{subfigure}
~
    \centering
    \begin{subfigure}[t]{0.3\textwidth}
        \centering
\includegraphics[width=\textwidth]{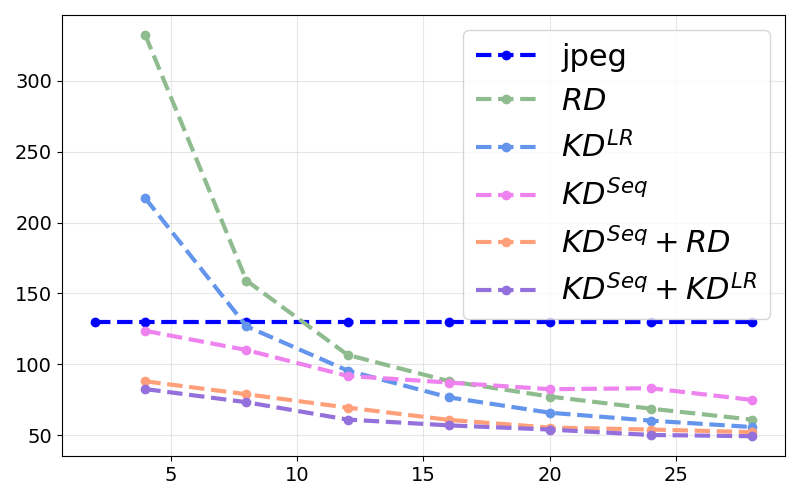}
    \end{subfigure}
\\
    \centering
    \begin{subfigure}[t]{0.3\textwidth}
        \centering
\includegraphics[width=\textwidth]{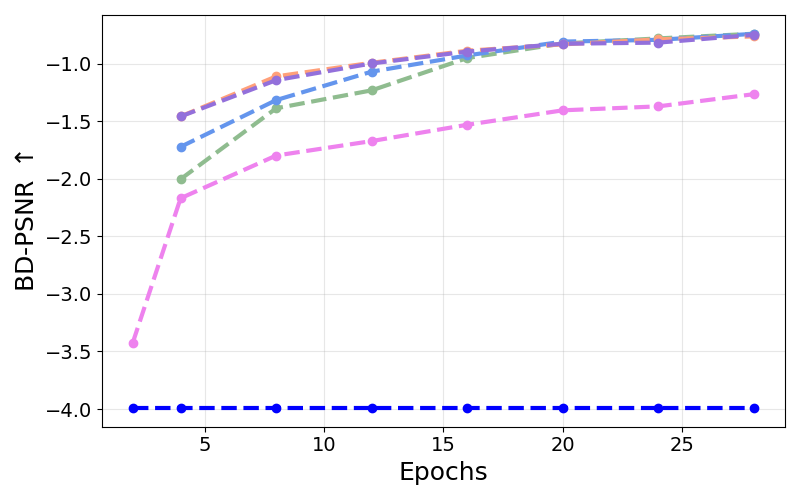}
    \caption{$\div 2$}
    \end{subfigure}
~
 \begin{subfigure}[t]{0.3\textwidth}
        \centering
\includegraphics[width=\textwidth]{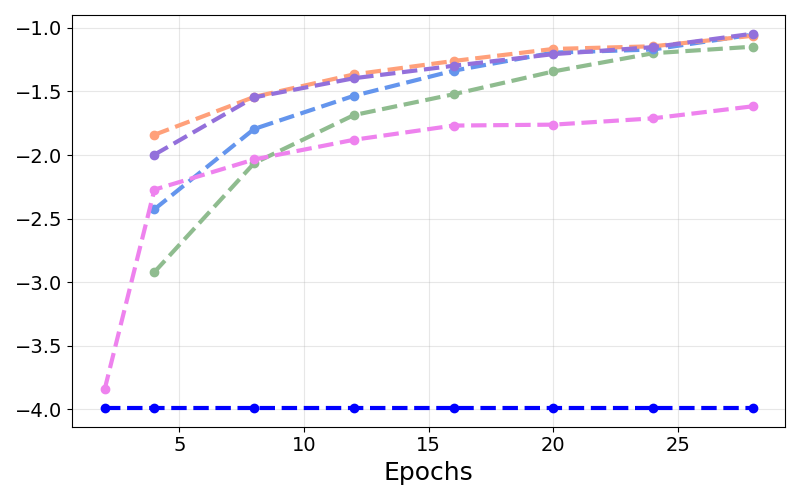}
 \caption{$\div 4$}
    \end{subfigure}
~
    \begin{subfigure}[t]{0.3\textwidth}
        \centering
\includegraphics[width=\textwidth]{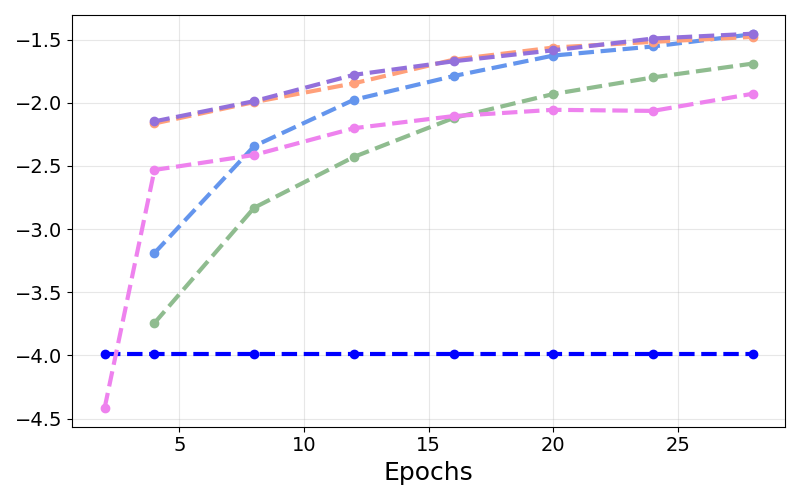}
 \caption{$\div 8$}
    \end{subfigure}
\caption{\small{\textit{After reduction, KD strategies --- especially those using $KD^{Seq}$ --- outperform $RD$ in early epochs.} BD-rate results in the first row, and BD-PSNR results in the second row (calculated with respect to the orginal full-size architecture), for $FP$ evaluated on CLIC2020 with reduction factors $r \in \{2, 4, 8\}$.}}
\label{fig:fact_clic}
\end{figure*}

\section{Experiments and results}

We selected two autoencoder architectures: the $FP$~\cite{balle:iclr:2017:factorized} and the $HP$-based MS-ILLM~\cite{muckley:2023:icml}. The former focuses on the reconstruction error $|x - \hat{x}|_2^2$ and is evaluated with PSNR, while the later incorporates perceptual metrics and a generative strategy to improve reconstructions, with evaluation including FID~\cite{mentzer:2020:neurips}. Implementations are based on CompressAI~\cite{begaint:2020:compressai} and NeuralCompression~\cite{muckley:2021:neuralcompression}, respectively. For $FP$, we used  $r \in \{2,4,8\}$, for MS-ILLM $r \in \{2,4\}$.  Training followed the standard protocols of CompressAI and NeuralCompression.

Trainings were conducted on Vimeo-90k triplet dataset with a total of $153,939$ images. For evaluation, we used 
the test split of CLIC2020~\cite{toderici:2020:clic} (428 images). Performance measured with BD-rate and BD-PSNR per epoch for the $FP$, and bit-rate vs. FID and bit-rate vs. PSNR for MS-ILLM. Code available upon acceptance. We compare the following architectures.

\textbf{Teachers ($Orig$):} We used 5 pre-trained models with different compression rates for $FP$, and 5 pre-trained models for MS-ILLM (trained using GANs to increase statistical fidelity). Teacher $FP$ models were trained on the Vimeo-90k dataset~\cite{xue:ijvc:2019:vimeo}, containing  89,800 video clips. Teacher MS-ILLM models were trained on OpenImages V6~\cite{Kuznetsova:2020:ijcv}.

\textbf{Original loss ($RD$):} We train models using the teachers' reduced architecture (encoders and decoders $\div r$)  \emph{from scratch} with the original loss functions (Equations~\eqref{eq:rd_loss} and \eqref{eq:rd_loss_hyper}).

\textbf{Latent regularization ($KD^{LR}$):} We train models using the teachers' reduced architecture (encoders and decoders $\div r$) with the original loss functions and an additional latent regularization term. Final loss functional is presented in Equation~\eqref{eq:rd_loss_regularized}.

\textbf{Sequential ($KD^{Seq}$):} We perform \emph{only the initialization stage} of our strategy (step $i)$ and $ii)$), that is, sequential training of the reduced encoder then decoder. As described above, this use the loss functions in Equations~\eqref{eq:loss_KD_alone}~and~\eqref{eq:loss_KD_alone_dec} for $FP$ and in Equations~\eqref{eq:loss_hyper_KD_alone}~and~\eqref{eq:loss_hyper_KD_alone_dec} for MS-ILLM. The hyper encoder and hyper decoder, on MS-ILLM, $h_{a,\mathcal{S}}(\cdot)$ and $h_{s,\mathcal{S}}(\cdot)$, were also reduced on the same rate $r$ as $g_{a,\mathcal{S}}(\cdot)$ and $g_{s,\mathcal{S}}(\cdot)$.

\textbf{Sequential + original loss ($KD^{Seq}+RD$):} We use the same training strategy as in \textit{Sequential} ($KD^{Seq}$) for \emph{two initial epochs}, after which we use the learned weights to continue training with the \textit{Original loss} ($RD$) strategy.

\textbf{Sequential + latent regularization ($KD^{Seq}+KD^{LR}$):} We use the training strategy as in \textit{Sequential} ($KD^{Seq}$) for \emph{two initial epochs}, after which we continue training with the \textit{Latent regularization} ($KD^{LR}$) strategy.

\begin{figure}[!ht]
    \centering
    \begin{subfigure}[t]{0.2\textwidth}
        \centering
\includegraphics[width=\textwidth]{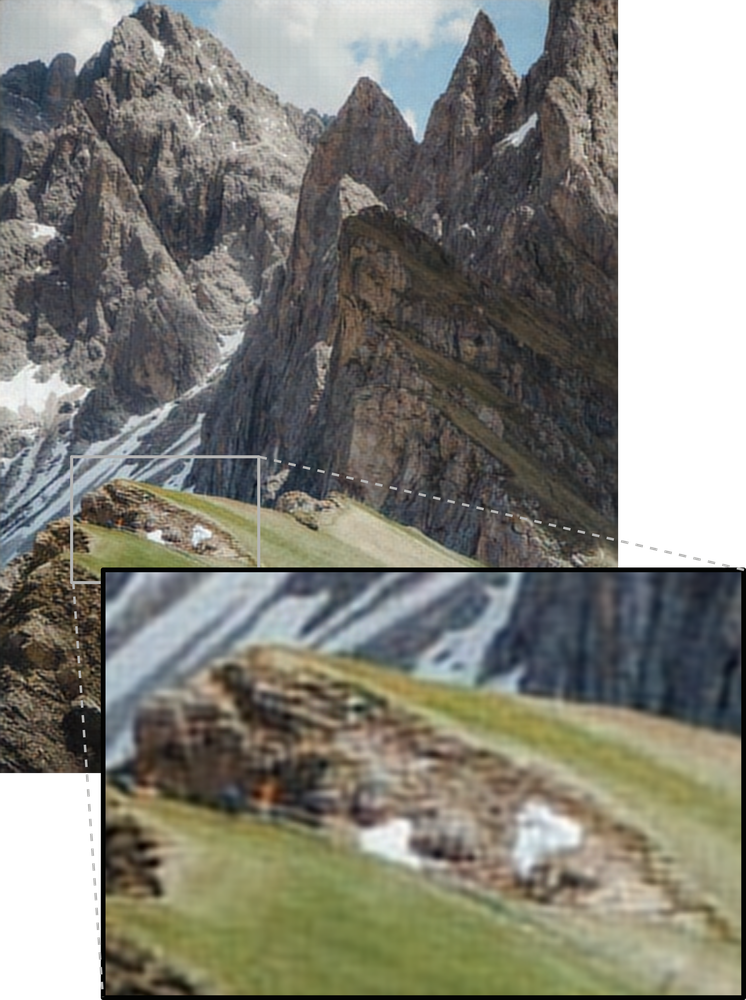}
    \caption{$RD$}
    \end{subfigure}
 ~
    \centering
    \begin{subfigure}[t]{0.2\textwidth}
        \centering
\includegraphics[width=\textwidth]{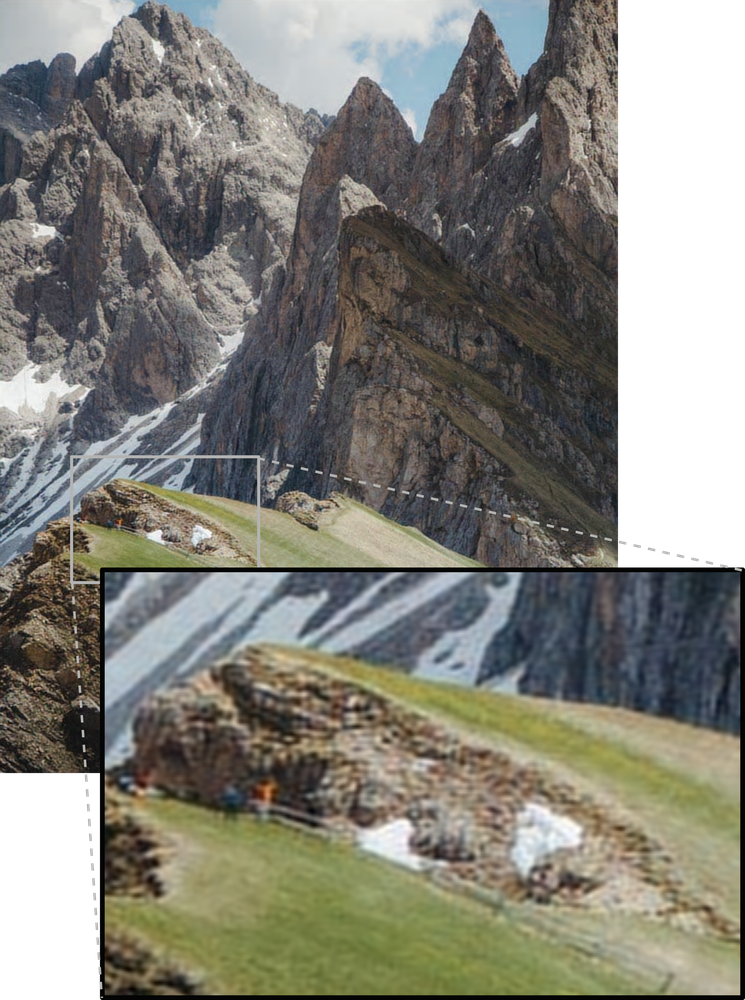}
    \caption{$KD^{Seq}+KD^{LR}$}
    \end{subfigure}
\caption{\small{\textit{$KD^{Seq} + KD^{LR}$ produces higher-quality images after few epochs}. Examples from $FP$ after 4 epochs with 0.5 of bit-rate. 
}}
\label{fig:example_images}
\end{figure}

\textbf{Image quality:} We show results for $FP$ using the Bjøntegaard-Delta (BD) Rate and BD-PSNR in Fig.~\ref{fig:fact_clic} for $FP$. The first row reports BD-Rate, for which lower values are better, while the second row reports BD-PSNR, where higher values indicate better performance. Each point on the curves corresponds to a training epoch. The proposed sequential $KD^{Seq}$ strategy outperforms $RD$ in early epochs; however, its performance saturates at later stages of training, that could indicate the need of introducing more complexity at this stage. This saturation is mitigated by the two-step strategy (sequential training followed by the original loss, with or without latent regularization), which consistently improves both rate and PSNR compared to training with the original loss alone. The best results are achieved by $KD^{Seq}+KD^{LR}$, followed by $KD^{Seq}+RD$, particularly at higher reduction levels (Fig.~\ref{fig:fact_clic}~(c)). This highlights the positive impact of KD in early stages of training, especially when using a simplified optimization objective ($KD^{Seq}$).
We present reconstructed images in Fig.~\ref{fig:example_images}, using $FP$ models with $r=8$ in epoch 4. The results are consistent with Fig.~\ref{fig:fact_clic}, showing that $KD^{Seq} + KD^{LR}$ models produce better visual reconstructions. 

\begin{figure*}[!ht]
\centering
\begin{subfigure}[t]{\textwidth}
\centering
\includegraphics[width=0.7\textwidth]{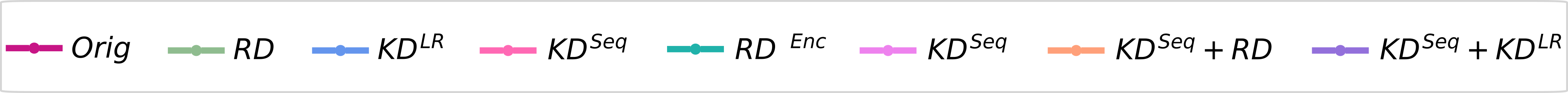}
    \end{subfigure}
\\
 \begin{subfigure}[t]{0.23\textwidth}
\centering
\includegraphics[width=\textwidth]{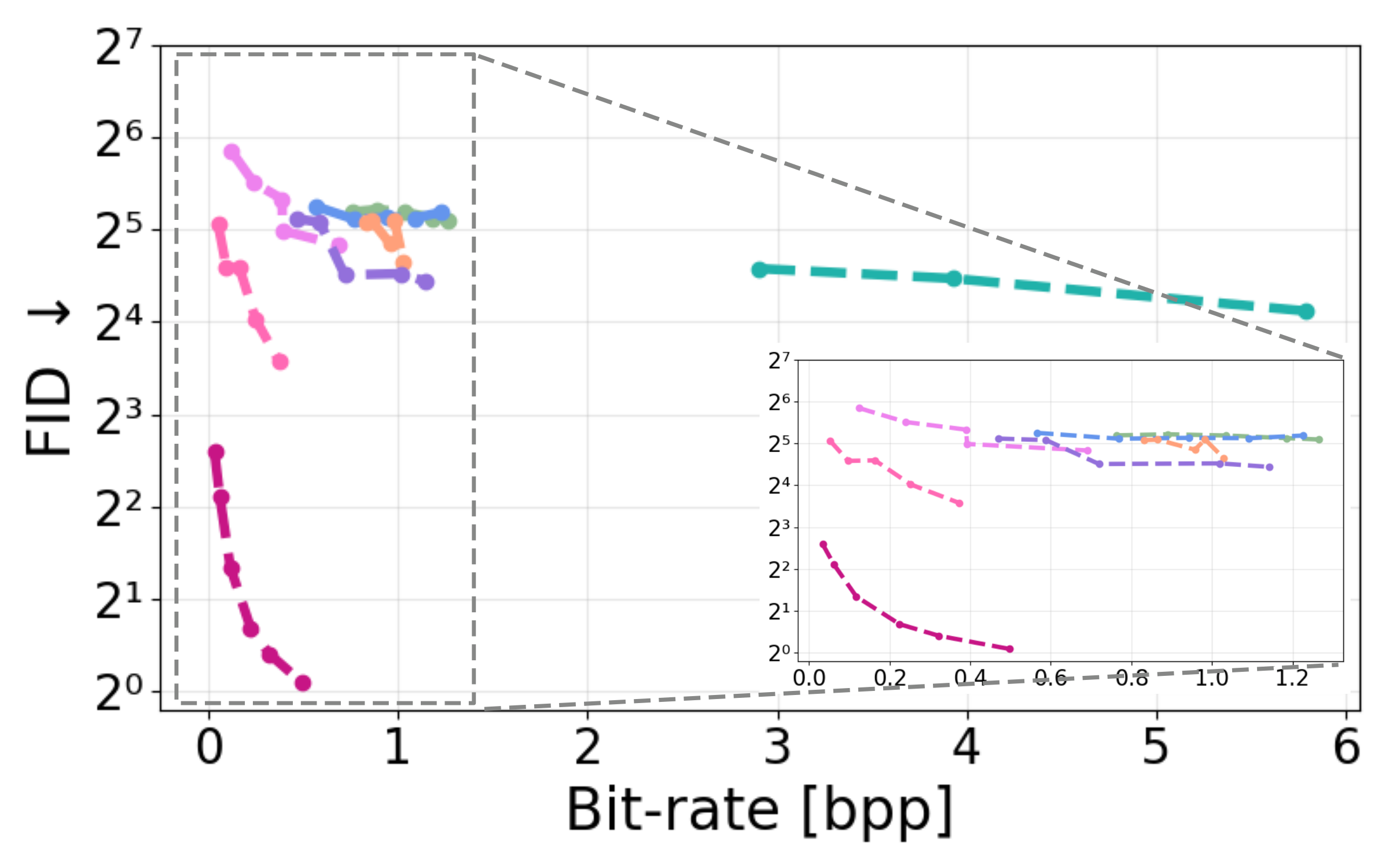}
 \caption{4 epochs}
    \end{subfigure}
~
    \begin{subfigure}[t]{0.23\textwidth}
        \centering
\includegraphics[width=\textwidth]{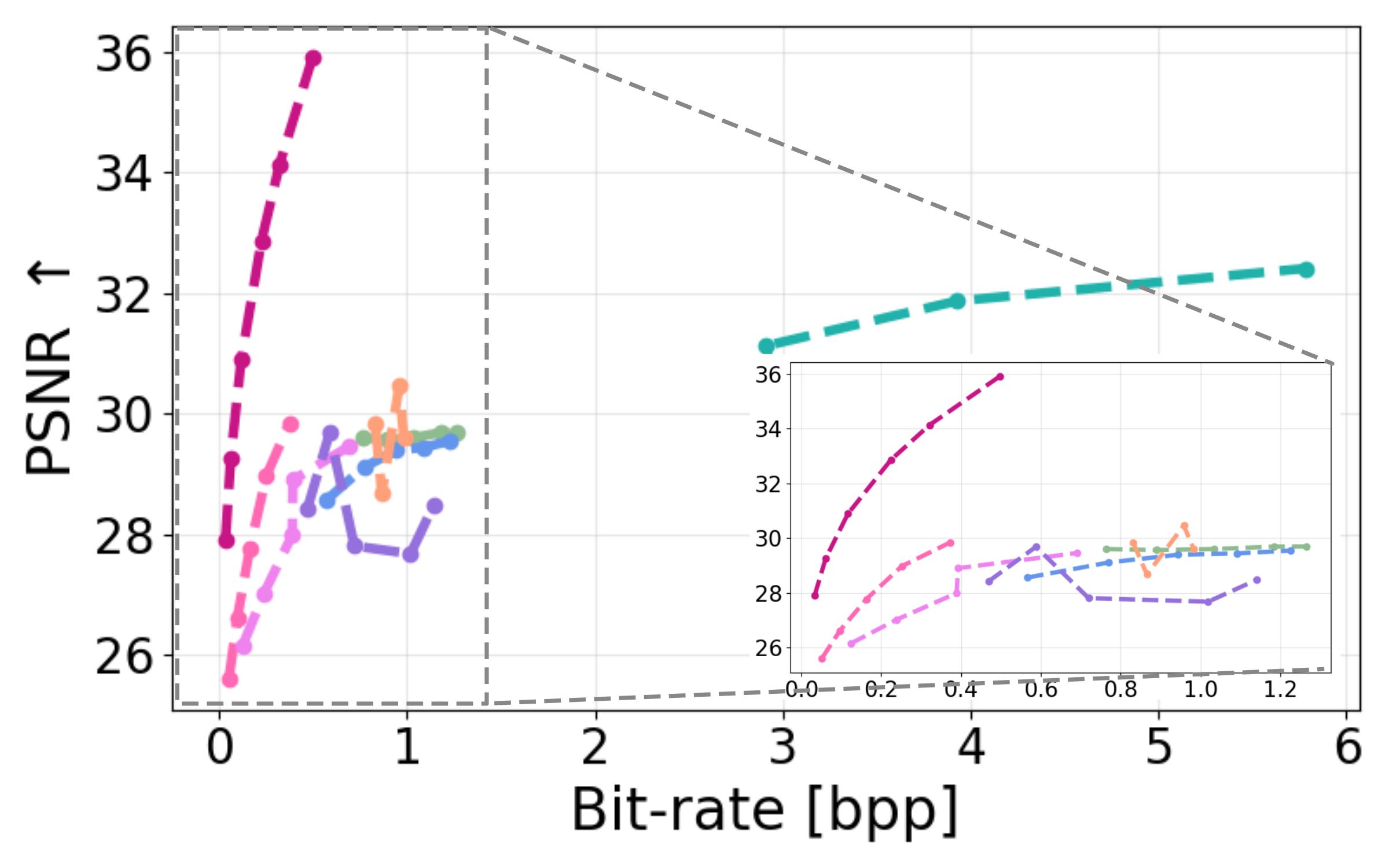}
\caption{4 epochs}
    \end{subfigure}
    ~
    \begin{subfigure}[t]{0.23\textwidth}
    \includegraphics[width=\textwidth]{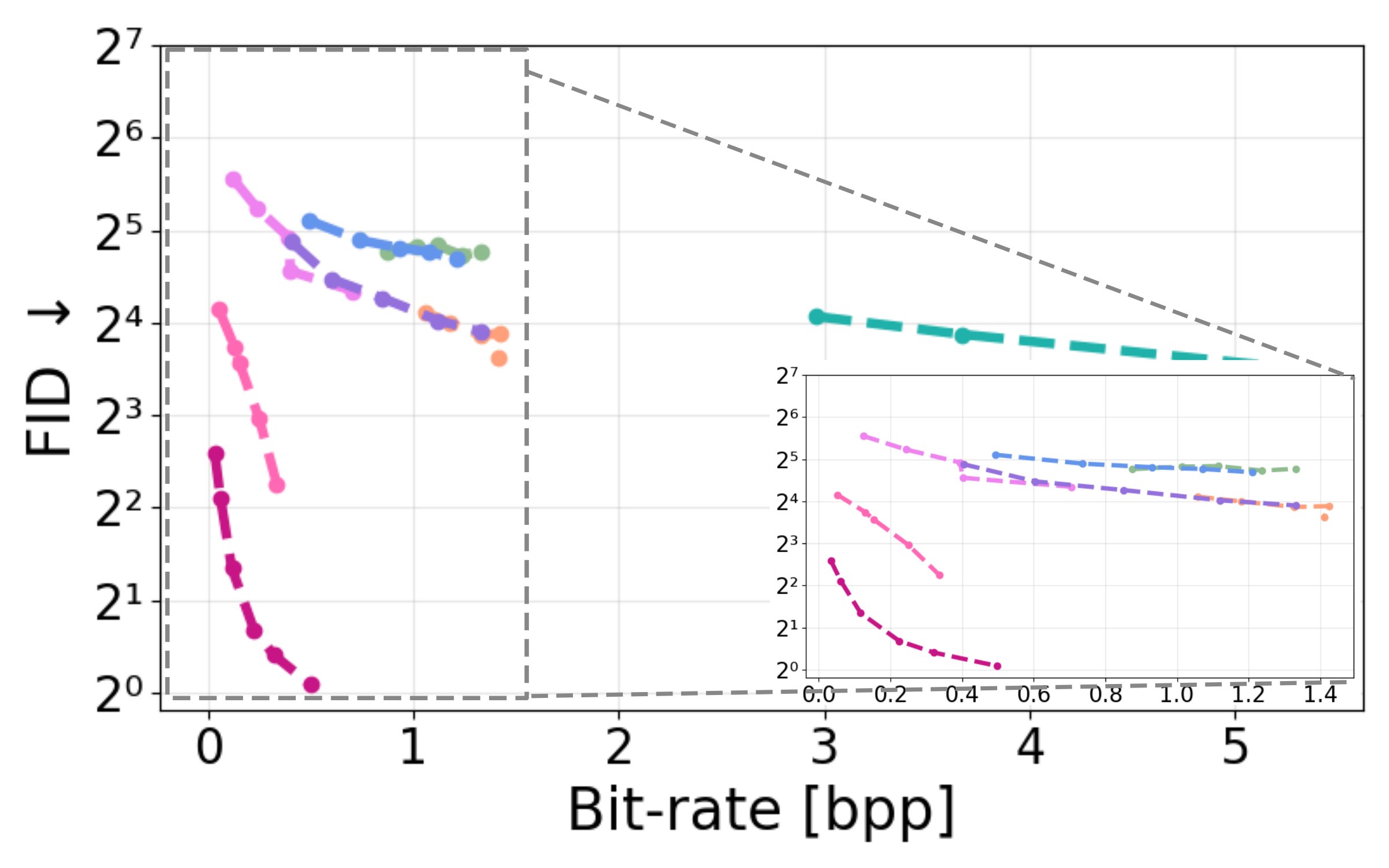}
    \caption{12 epochs}
    \end{subfigure}
~
    \begin{subfigure}[t]{0.23\textwidth}
        \centering
\includegraphics[width=\textwidth]{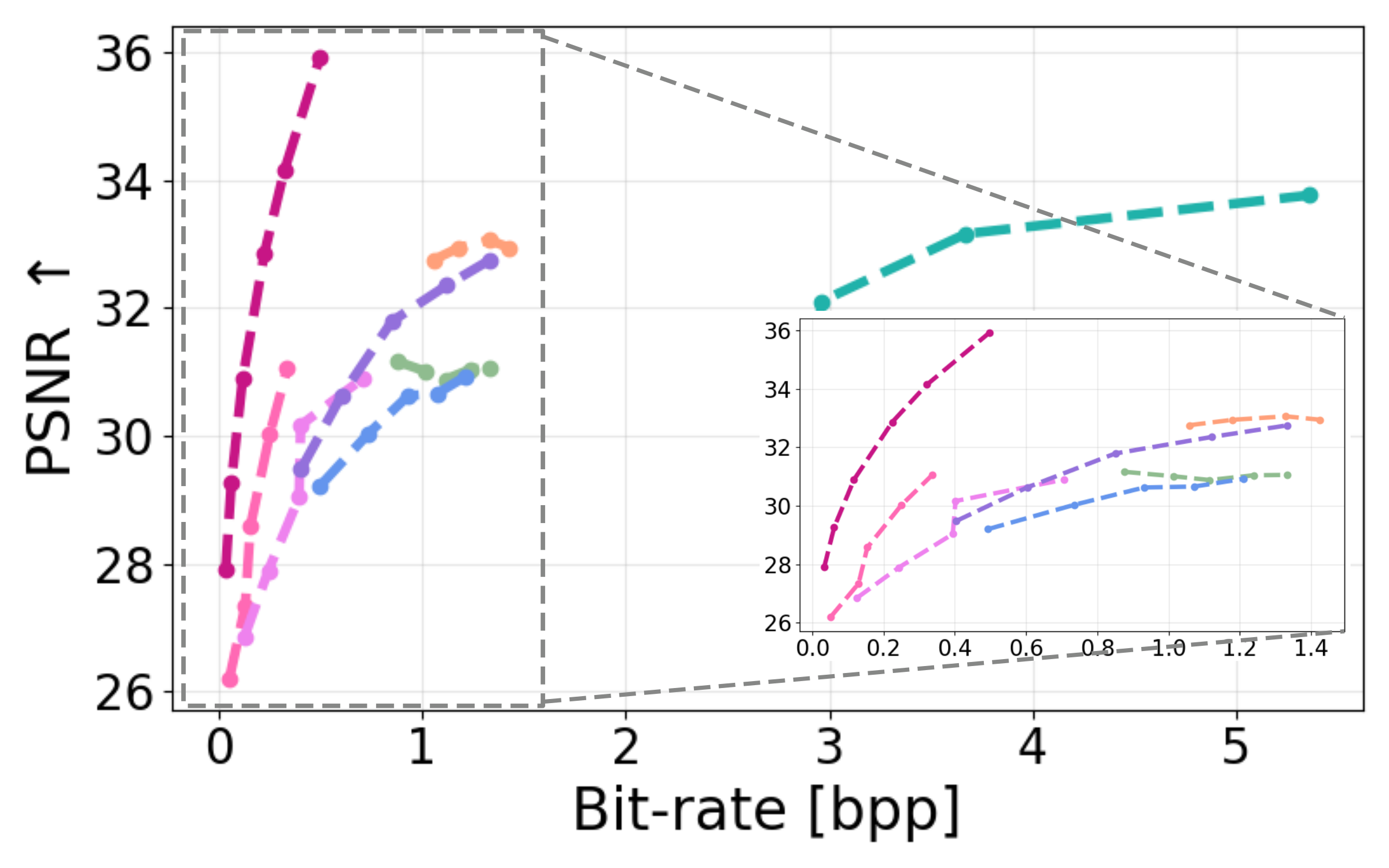}
\caption{12 epochs}
    \end{subfigure}
\caption{\small{\textit{After reduction, KD strategies --- especially those using $KD^{Enc}$ --- outperform $RD$ in rate-distortion and fidelity evaluation.} Bit-rate vs. FID results in (a) and (c), and Bit-rate vs. PSNR  in (b) and (d), for $HP$ evaluated on CLIC2020 with reduction factors $r \in \{4\}$ (more aggressive reduction) in epochs 4 and 12. $RD^{Enc}$ produced bit-rates higher than 3.0, which distorted the visualization of the other curves. Therefore, we provide zoomed-in graphs. Since the models aim to improve statistical fidelity, we do not compare here with jpeg.}}
\label{fig:illm_clic}
\end{figure*}

Fig.~\ref{fig:illm_clic} presents the Bit-rate vs. FID and Bit-rate vs. PSNR results at epochs 4 and 12 during the training of the complex architecture (MS-ILLM) based on \textit{HP}. One can see that the reduction does not achieve satisfying results yet, as the performance of all reduced architectures is still far from the original scheme in terms of rate-PSNR and rate-FID. One can however, remark that the  $KD^{Seq}$ strategy still outperforms the other reduction strategies, highlighting the potential of relying on our proposed sequential knowledge distillation.

The reason why the reduction of MS-ILLM is not as easy as the \textit{FP} scheme is that MS-ILLM relies on a GAN during part of the training. In order to highlight the benefits of our KD approach on a \textit{HP}-based architecture, we propose to consider only the reduction of the encoder. More precisely, Fig.~\ref{fig:illm_clic} also depicts the performance of the following two methods:

\textbf{Encoder + original loss ($RD^{Enc}$):} We train models with reduced encoders ($\div r$)  \emph{from scratch} with the original loss function (Equation\eqref{eq:rd_loss_hyper}), and $g_s^{\mathcal{S}}(\cdot) = g_s^{\mathcal{T}}(\cdot)$ and $h_s^{\mathcal{S}}(\cdot) = h_s^{\mathcal{T}}(\cdot)$.

\textbf{Encoder ($KD^{Enc}$):} We perform only the encoder reduction (step $i)$). This is made by using Equation~\eqref{eq:loss_hyper_KD_alone} as the loss function. The hyper encoder $h_{a,\mathcal{S}}(\cdot)$ is also reduced on the same rate $r$ as $g_{a,\mathcal{S}}(\cdot)$, and $g_s^{\mathcal{S}}(\cdot) = g_s^{\mathcal{T}}(\cdot)$ and $h_s^{\mathcal{S}}(\cdot) = h_s^{\mathcal{T}}(\cdot)$.

\begin{figure}[!ht]
    \centering
    \begin{subfigure}[t]{0.2\textwidth}
        \centering
\includegraphics[width=\textwidth]{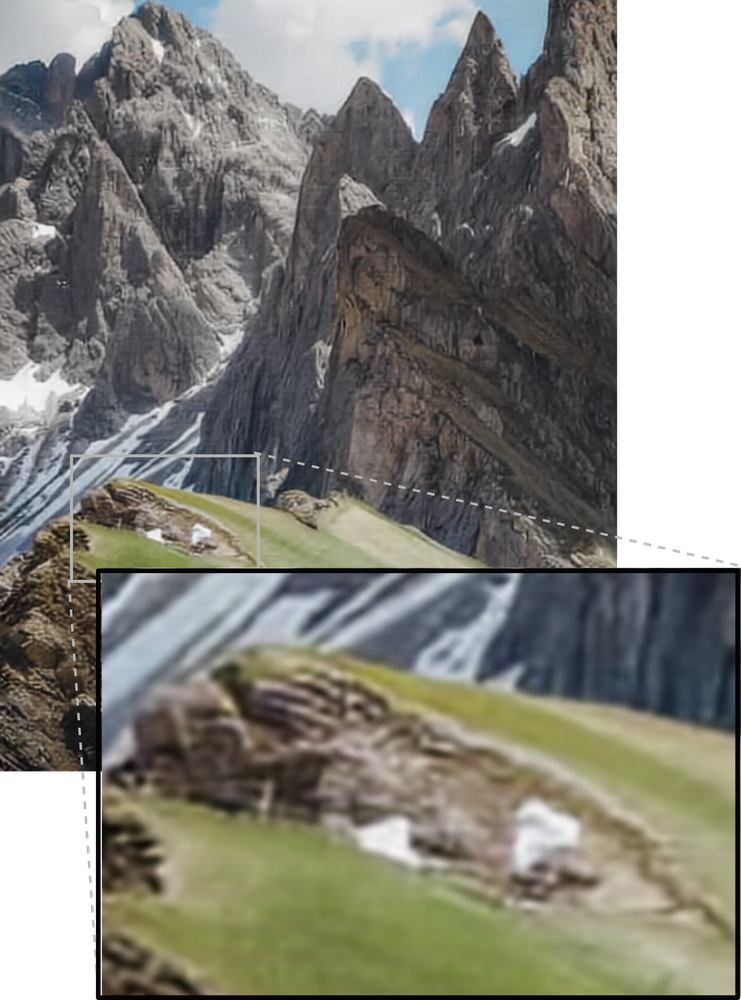}
    \caption{$RD^{Enc}$}
    \end{subfigure}
 ~
    \centering
    \begin{subfigure}[t]{0.2\textwidth}
        \centering
\includegraphics[width=\textwidth]{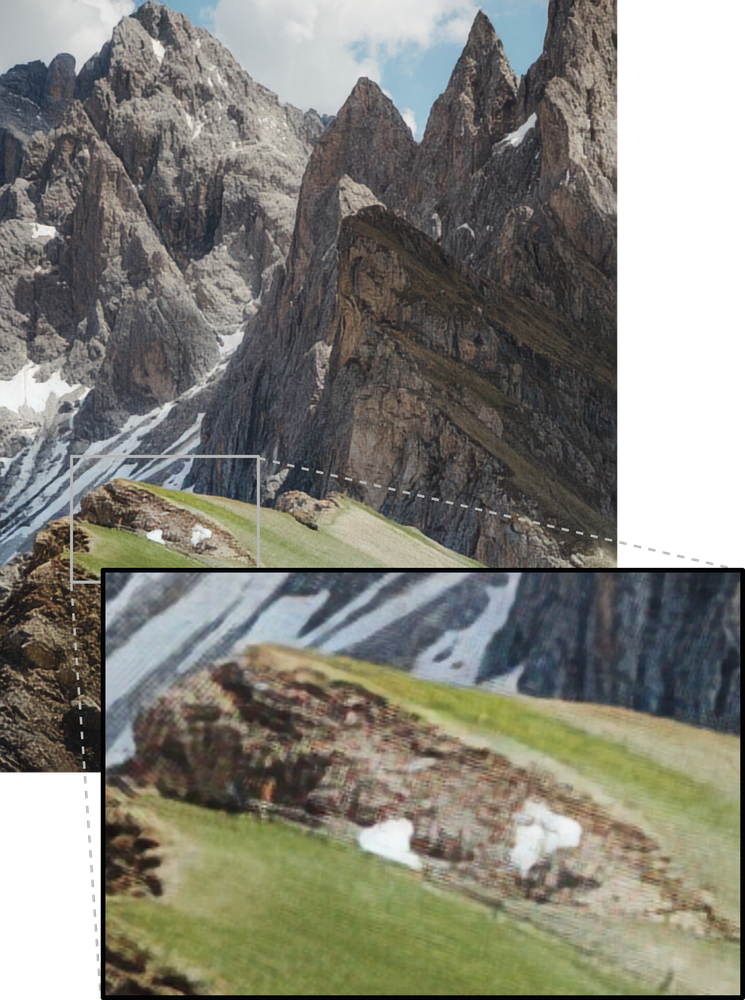}
    \caption{$KD^{Enc}$}
    \end{subfigure}
\caption{\small{\textit{$KD^{Enc}$ produces higher-quality images after few epochs}. Examples from $HP$ after 4 epochs. $RD^{Enc}$ has bit-rate more than 7 times higher.}}
\label{fig:example_images_illm}
\end{figure}

We can see that $KD^{Enc}$ achieves much better performance than $RD^{Enc}$. This can be explained by the fact that the proposed latent-based loss avoids using a GAN while keeping its benefits. The absolute performance of the proposed reduced encoders becomes interesting and closer to the original. We present reconstructed images in Fig.~\ref{fig:example_images_illm} to support these remarks.
It shows that the proposed knowledge distillation is promising for the reduction of an \textit{HP}-based encoder. Future work may focus on a reliable reduction of the decoder. More precisely, future studies should investigate: (i) analyzing additional epochs as possible transition points in the combined $KD^{Seq} + KD^{LR}$ training strategy, and (ii) exploring alternative decoder reduction strategies, particularly for models based on generative behavior for reconstruction.

\textbf{Reduction ratio:} Table~\ref{tab:compression} shows the reduction in operations and models' storage for different reduction rates on the width (number of filters) across the layers. The highest reduction rate ($\div  8$) achieves a reduction of over $35\times$ in Multiply-Accumulate Operations (MACs) and $16\times$ in the architecture storage, compared to the original architecture ($\div 1$) for $FP$.

\begin{table}[!h]
\centering
\caption{\small{\textit{Compressing autoencoders leads to significant reductions on MACs.} Multiply-Accumulate Operations (MACs) for architecture width reductions, normalized to the full-width ($\div 1$), 194,903.5 MACs/pixel for the $FP$ and 975,865.7 MACs/pixel for MS-ILLM.}}
\label{tab:compression}
\resizebox{9cm}{!} 
{ 
\begin{tabular}{cc|rrrr}
\multicolumn{2}{c|}{\textbf{Width reduction}}                                                    & \multicolumn{1}{c}{$\div 8$} & \multicolumn{1}{c}{$\div 4$} & \multicolumn{1}{c}{$\div 2$} & \multicolumn{1}{c}{$\div 1$} \\ \hline
\multicolumn{1}{c|}{\multirow{2}{*}{\textbf{Relative MACs}}}               & \textbf{Factorized} & 0.029                       & 0.086                       & 0.281                       & 1.000                       \\
\multicolumn{1}{c|}{}                                                      & \textbf{MS-ILLM}    & -                            & 0.071                       & 0.261                       & 1.000                       \\ \hline
\multicolumn{1}{c|}{\multirow{2}{*}{\textbf{Size weights (MB)}}} & \textbf{Factorized} & 0.8                       & 1.8                       & 4.3                       & 12.0                       \\
\multicolumn{1}{c|}{}                                                      & \textbf{MS-ILLM}    & -                            & 55.6                       & 195.1                      & 726.0                     
\end{tabular}
}
\end{table}



\section{Conclusion}

In this paper, we propose a methodology that adapts \textit{Knowledge Distillation} (KD) to reduce image compression autoencoders within a limited number of training epochs. The main idea is to simplify the training loss and perform a sequential reduction to avoid overly aggressive architectural compression. We show that, under constraints on model size and training time, our method outperforms naive architecture reduction, achieving improved quantitative and qualitative results. Furthermore, when applied to complex teacher models that incorporate generative strategies, such as MS-ILLM, our approach leads to more effective training initialization. 


\bibliographystyle{IEEEtrans}
\bibliography{IEEEabrv,main}

\end{document}